\pdfoutput=1

\documentclass[11pt]{article}

\usepackage{acl}

\usepackage{times}
\usepackage{latexsym}

\usepackage[T1]{fontenc}

\usepackage[utf8]{inputenc}

\usepackage{microtype}

\usepackage{inconsolata}

\usepackage{graphicx}
\usepackage{multirow}
\usepackage{booktabs}

\usepackage{algorithm}
\usepackage{algpseudocode}
\usepackage{adjustbox}
\usepackage{amsmath}
\usepackage{amssymb}
\usepackage{cleveref
}
\title{Question Decomposition for Retrieval-Augmented Generation}

\author{
    Paul J. L. Ammann \hspace*{15mm} Jonas Golde \hspace*{15mm} Alan Akbik \vspace*{2mm}\\
    Humboldt-Universität zu Berlin \\
    \texttt{\{paul.ammann, jonas.max.golde.1, alan.akbik\}@hu-berlin.de}
}

\begin{document}
\maketitle
\begin{abstract}
Grounding large language models (LLMs) in verifiable external sources is a well-established strategy for generating reliable answers. Retrieval-augmented generation (RAG) is one such approach, particularly effective for tasks like question answering: it retrieves passages that are semantically related to the question and then conditions the model on this evidence. However, multi-hop questions, such as \textit{``Which company among NVIDIA, Apple, and Google made the biggest profit in 2023?,''} challenge RAG because relevant facts are often distributed across multiple documents rather than co-occurring in one source, making it difficult for standard RAG to retrieve sufficient information. To address this, we propose a RAG pipeline that incorporates question decomposition: (i) an LLM decomposes the original query into sub-questions, (ii) passages are retrieved for each sub-question, and (iii) the merged candidate pool is reranked to improve the coverage and precision of the retrieved evidence. We show that question decomposition effectively assembles complementary documents, while reranking reduces noise and promotes the most relevant passages before answer generation. Although reranking itself is standard, we show that pairing an off-the-shelf cross-encoder reranker with LLM-driven question decomposition bridges the retrieval gap on multi-hop questions and provides a practical, drop-in enhancement, without any extra training or specialized indexing. We evaluate our approach on the MultiHop-RAG and HotpotQA, showing gains in retrieval ($MRR@10: +36.7\%$) and answer accuracy ($F1: +11.6\%$) over standard RAG baselines.

\end{abstract}

\label{sec:overview}
\begin{figure}[ht]
\includegraphics[width=\linewidth]{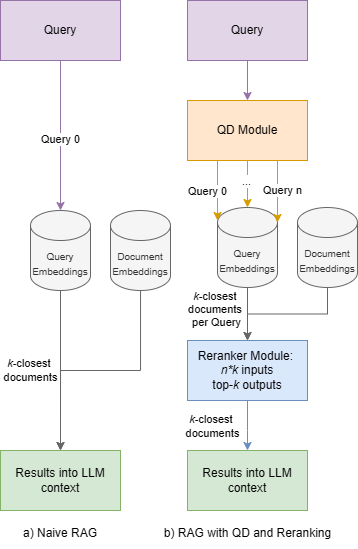}
\caption{(a) Standard retrieval in RAG versus (b) our approach using question decomposition and reranking.}
\label{figure:naive_rag_vs_qd}
\end{figure}

\section{Introduction}

Retrieval-augmented generation (RAG) addresses knowledge gaps in large language models (LLMs) by retrieving external information at inference time \citep{lewisRetrievalAugmentedGenerationKnowledgeIntensive2020}. While effective, RAG's performance depends heavily on retrieval quality; irrelevant documents can mislead the model and degrade the quality of its output \citep{choImprovingZeroshotReader2023, shiLargeLanguageModels2023}. For example, when asked ``Who painted \textit{Starry Night}?'' a naive retriever may surface a general Wikipedia article on \emph{Post-Impressionism} rather than the specific page on \emph{Vincent van Gogh}, offering little direct evidence for the correct answer. This issue becomes more pronounced in multi-hop QA tasks, where supporting facts are spread across multiple documents. For instance, a single, undifferentiated search for the query ``Which company among NVIDIA, Apple, and Google made the biggest profit in 2023?'' might return a broad market overview article mentioning all three companies together, but omit their individual 2023 earnings reports—forcing the model to respond without access to the necessary disaggregated information.

\noindent\textbf{Challenges of Multi-hop Retrieval.} Complex questions often require reasoning over multiple entities, events, or steps, which are rarely addressed within a single document. While the individual facts needed to answer such questions may be simple, the required evidence is typically distributed across multiple sources. To improve retrieval coverage in multi-hop QA settings, our approach decomposes the original question into simpler subqueries---a process we refer to as \emph{question decomposition} \citep{perezUnsupervisedQuestionDecomposition2020a}. By breaking down a complex query into focused subqueries, question decomposition increases the likelihood of retrieving documents that address distinct aspects of the information need, especially when information sources are self-contained.

Consider the question: ``Which planet has more moons, Mars or Venus?'' In a standard RAG pipeline, the entire question is embedded as a single unit, and the retriever attempts to find a single passage that answers it directly (cf.~\Cref{figure:naive_rag_vs_qd}a). In practice, this often results in retrieving a general article about planetary science or solar system formation. We assume that relevant facts are located in two self-contained documents—one about Mars and the other about Venus. With QD, we exploit the fact of increasingly capable LLMs to generate fact-seeking subquestions such as ``How many moons does Mars have?'' and ``How many moons does Venus have?'', each of which is more likely to retrieve a precise, relevant answer from its respective source (cf.~\Cref{figure:naive_rag_vs_qd}b).

\noindent\textbf{Contributions.} In this paper, we present a retrieval-augmented generation pipeline that integrates question decomposition with reranking to improve multi-hop question answering. Our QD component uses a LLM to decompose complex questions into simpler subqueries, each addressing a specific part of the information need, and thus requires no fine-tuning or task-specific training. Retrieved results from all subqueries are aggregated to form a broader and more semantically relevant candidate pool.

To mitigate the noise introduced by retrieving documents for each subquery, we apply a pre-trained reranker that scores each candidate passage based on its relevance to the original complex query. This substantially improves precision by filtering out irrelevant results. In combination, question decomposition ensures broad evidence coverage, while reranking distills this expanded set into a concise collection of highly relevant passages. 

We evaluate our approach on the MultiHop-RAG and HotpotQA benchmarks and demonstrate substantial gains in recall and ranking metrics over standard RAG and single-component variants. We further analyze the inference overhead, showing that the added cost of QD remains manageable. Our main contributions are as follows:

\begin{itemize}
    \item We propose a question decomposition (QD)–based RAG pipeline for multi-hop question answering, where a LLM decomposes complex questions into simpler subqueries without any task-specific training.
    \item To improve precision, we incorporate a cross-encoder reranker that scores retrieved passages based on their relevance to the original complex query, effectively filtering noise from the expanded candidate pool introduced by QD.
    \item We empirically validate our approach on the MultiHop-RAG and HotpotQA benchmarks, demonstrating substantial improvements in retrieval recall, ranking quality, and final answer accuracy—achieved without any domain-specific fine-tuning.
\end{itemize}
We release our code\footnote{https://github.com/Wecoreator/qd\_rag} on GitHub for reproducibility.

\section{Methodology} \label{sec:methodology}

Our pipeline follows the retrieval-augmented generation framework of \citet{lewisRetrievalAugmentedGenerationKnowledgeIntensive2020}, which combines a retriever with a generative language model. The goal is to answer a natural language query $q$ by grounding the language model's response in documents retrieved from a large corpus $\mathcal{D}$.


\paragraph{Retrieval.}
In the first step, a query encoder $f_q$ and a document encoder $f_d$ project queries and documents into a shared vector space \citep{karpukhinDensePassageRetrieval2020}. During retrieval, the query representation $f_q(q)$ is compared to all document embeddings $f_d(d)$ using inner product similarity. Subsequently, we select the top-$k$ most relevant documents:

\[
R(q) = \mathrm{Top}\text{-k}_{d \in \mathcal{D}} \left( \langle f_q(q), f_d(d) \rangle \right)
\]

Here, $\langle \cdot,\cdot \rangle$ denotes the similarity score between the query and document embeddings, computed as inner product similarity in the shared embedding space. This dense retrieval stage identifies documents that are semantically similar to the query and provides candidates for grounding the language model's response.

\paragraph{Reranking.}
To refine the initial retrieval set $R(q)$, we apply a pre-trained reranker that computes fine-grained relevance scores between the query $q$ and each candidate document $d \in R(q)$. Cross-encoder rerankers are a staple of modern information retrieval and already feature in recent RAG systems \citep{glassRe2GRetrieveRerank2022, wangSearchingBestPractices2024a}. We therefore deliberately employ an off-the-shelf model. Each query-document pair is jointly encoded by a transformer model, producing a single relevance score $g_{\phi}(q, d) \in \mathbb{R}$, where $\phi$ denotes the model parameters. The top-$k$ documents (ranked in descending order of $g_{\phi}(q, d)$) form the final reranked set $R'(q)$. Only these top-$k$ ranked passages are passed to the generator, while the rest are discarded. Unlike the retrieval stage, where queries and documents are encoded independently for efficiency, reranking involves joint encoding of each pair, which increases computational cost but enables more accurate relevance estimation by modeling interactions between query and document tokens.


\paragraph{Generation.}
A pretrained autoregressive LLM receives the concatenation of $q$ and the top-ranked passages and then generates the answer. Specifically, we concatenate the query with the top-ranked passages $R'(q) = \{d_1, \dots, d_r\}$ into a single input sequence:

\[
x = [q; d_1; d_2; \dots; d_r]
\]

The model then generates the answer token-by-token, modeling the conditional probability:

\[
p(y \mid x) = \prod_{t=1}^{T} p(y_t \mid y_{<t}, x).
\]

This way, we enable the language model to attend over the complete retrieved context and generate a response grounded in multiple evidences simultaneously.




\section{RAG with Question Decomposition}


A \emph{naive} RAG system encodes the user query $q$ once and retrieves the top-$k$ most relevant passages. These retrieved documents are then concatenated with the query and used as input to the language model, which generates an answer \citep{lewisRetrievalAugmentedGenerationKnowledgeIntensive2020,karpukhinDensePassageRetrieval2020}. Notably, this baseline assumes that the top-ranked passages contain all necessary evidence, treating each question as single-hop and ignoring multi-step reasoning or dependencies across documents.

Our proposed pipeline augments the standard RAG framework with two additional components: a \emph{question decomposition} module and a \emph{reranking} module. A comparison between our approach and a naive RAG baseline is illustrated in~\Cref{figure:naive_rag_vs_qd}. To address the challenges posed by multi-hop questions, which can degrade retrieval performance in standard RAG, we (i) decompose the original query into a set of simpler sub-queries, (ii) retrieve documents for each sub-query, (iii) merge and deduplicate the retrieved results, and (iv) apply a reranker to filter out noisy or weakly relevant candidates. From this filtered set, only the top-$k$ passages $R'(q)$ are passed to the language model. The full pipeline is described in~\Cref{alg:qd}.

\begin{algorithm*}[t]
\label{alg:qd}
\begin{algorithmic}[1]
\Require Query $q$, documents $\mathcal{D}$, cutoff $k$
\Ensure $R'(q)$: top-$k$ passages relevant to $q$
\State $Q \gets \{\,q\,\} \cup\textsc{Decompose}(q_0)$ \Comment{original and decomposed queries}
\State $C \gets \emptyset$\Comment{global candidate set}
\ForAll{$q \in Q$}
    \State $C \gets C \cup \textsc{Top\text{-}k}(q,\mathcal{D})$ \Comment{Add top-k candidates for each query}
\EndFor
\State $C \gets \textsc{Rerank}(C)$\Comment{using a pre-trained reranker}
\State $R'(q) \gets \textsc{Head}(C,k)$\Comment{retain highest-scoring $k$}
\State \Return $R'(q)$
\end{algorithmic}
\caption{Retrieval with question decomposition: Given a complex query $q$, the algorithm first generates sub-queries using an LLM, retrieves documents for each, and aggregates the results. A reranker then filters the merged candidate set, and the top-$k$ passages are selected for downstream generation.}
\end{algorithm*}


\subsection{QD Module}
Given a complex question \( q \), we define a prompting function \( \textsc{Decompose}(q, p) \) that produces a set of sub-queries \( \{\tilde{q}_1, \dots, \tilde{q}_n\} \), where \( p \) is a fixed natural language prompt provided to an instruction-tuned language model. The number of sub-queries \( n \) is not fixed but typically small, depending on how many distinct aspects or reasoning steps are involved in answering \( q \). The final set of queries used for retrieval is defined as \( Q = \{q\} \cup \textsc{Decompose}(q, p) \), where the original query \( q \) is always retained to preserve baseline retrieval performance.



\subsection{Reranker Module}
Decomposing a complex question $q$ into multiple sub-queries \( \{\tilde{q}_1, \dots, \tilde{q}_n\} \) naturally increases retrieval coverage but also introduces the risk of noise. Since documents are retrieved independently for each sub-query, some may be overly specific, only partially relevant, or even unrelated to the original question. To address this, we apply a reranking module that scores each retrieved document based on its relevance to the original complex query \( q \). This step helps to realign the expanded candidate pool with the user's initial intent by filtering out documents that, while relevant to a sub-question, do not meaningfully contribute to answering \( q \) as a whole. The goal is to retain only passages that clearly address distinct aspects of the original question, improving precision in the final evidence set.

\section{Experiments}
We evaluate our proposed question decomposition pipeline on established multi-hop question answering benchmarks, focusing specifically on the retrieval stage. This allows us to isolate and directly measure improvements in evidence selection, independent of downstream generation. Following prior work, we report results on the evaluation split, as gold test labels are not publicly available.

\subsection{Datasets}
We use the following datasets in our experiments:

\paragraph{MultiHop-RAG.} MultiHop-RAG \citep{tangMultiHopRAGBenchmarkingRetrievalAugmented2024} is specifically designed for RAG pipelines and requires aggregating evidence from multiple sources to answer each query. In addition to question-answer pairs, it provides gold evidence annotations, enabling fine-grained evaluation of both retrieval accuracy and multi-hop reasoning. Importantly, the retrieval and generation components are evaluated separately, allowing for focused analysis of each component. This separation allows fair comparison across systems.

\paragraph{HotpotQA.} HotpotQA \citep{yangHotpotQADatasetDiverse2018} is a widely used multi-hop question answering benchmark constructed over Wikipedia. It features questions that explicitly require reasoning over two or more supporting passages. Gold answers and annotated supporting facts are provided, making it suitable for evaluating both retrieval and end-to-end QA performance. In this work, we focus on retrieval accuracy to assess how well different strategies recover the necessary evidence.

\subsection{Baselines}



To assess the individual and combined contributions of question QD and reranking within multi-hop RAG, we evaluate four system configurations:

\begin{enumerate}
\item \textbf{Naive RAG} is the base setup in which a single query \( q \) is embedded, and the top-$k$ most relevant passages are retrieved from the corpus \( \mathcal{D} \) using dense retrieval.

\item \textbf{RAG + QD} modifies the retrieval stage by introducing question decomposition. The original query \( q \) is transformed into a set of sub-queries \( \{\tilde{q}_1, \dots, \tilde{q}_n\} \), and retrieval is performed independently for each element of \(Q = \{q\} \cup \{\tilde{q}_i\}\). The retrieved results are merged, and the top-$k$ passages are selected based on similarity scores. This setup increases retrieval coverage by capturing information across multiple query aspects.

\item \textbf{RAG + Reranker} retains the single-query retrieval approach but adds a reranking step. To support more diverse initial candidates, we retrieve the top-$2k$ passages for the original query ($2 \times k$ candidates), which are then scored by a reranker. The top-$k$ passages according to this score are selected as final input.

\item \textbf{RAG + QD + Reranker} combines both components. It first decomposes the query into sub-queries, retrieves documents for each, merges the results, and applies reranking to select the final top-$k$ passages. This configuration aims to improve both evidence coverage and ranking precision in multi-hop QA scenarios.
\end{enumerate}

\subsection{Evaluation Metrics}
\label{sec:evaluation-metrics}

We report dataset-specific evaluation metrics in accordance with the protocols defined for each benchmark.

\paragraph{MultiHop-RAG.}
Following \citet{tangMultiHopRAGBenchmarkingRetrievalAugmented2024}, we report the following three retrieval-oriented metrics:
\begin{itemize}
    \item \textbf{Hits@$k$} for $k\!\in\!\{4,10\}$ which represents the percentage of questions for which at least one gold evidence appears in the top-$k$ retrieved passages.
    \item \textbf{MAP@10} (\emph{mean average precision}) computes the average precision at each rank position where a gold passage is retrieved, and then averages this over all queries. We truncate at rank 10.
    \item \textbf{MRR@10} (\emph{mean reciprocal rank}) computes the mean of the reciprocal rank of the \textit{first} correct passage, rewarding systems that surface a gold document as early as possible. We also truncate at rank 10.
\end{itemize}

\paragraph{HotpotQA.}
For HotpotQA, we adopt the official QA-centric evaluation metrics introduced in the original benchmark \citep{yangHotpotQADatasetDiverse2018,rajpurkarSQuAD100000Questions2016}. Results are reported separately for (i) answer accuracy, (ii) supporting fact prediction, and (iii) their joint correctness. The joint metric constitutes a stricter criterion, requiring both the predicted answer and the corresponding supporting evidence to be correct. This provides a more comprehensive assessment of system performance by jointly evaluating generation quality and the relevance of retrieved evidence.

\begin{itemize}
    \item \textbf{EM} (\emph{exact match}) measures whether the predicted answer exactly matches the reference answer string.
    \item \textbf{F1, Precision, Recall} measure token-level overlap between the predicted and reference answers, thus allowing for partially correct answers.
    \item \textbf{Supporting-Fact EM, F1, Precision, Recall} are the same metrics applied to the gold-labeled supporting facts.
    \item \textbf{Joint EM, F1, Precision, Recall} considers a prediction correct only if both the answer and \emph{all} supporting facts are correct. This metric captures the system's ability to jointly generate correct answers and identify the correct supporting evidence.
\end{itemize}

\subsection{Implementation Details}
\paragraph{Retrieval} We embed each passage chunk using \texttt{bge-large-en-v1.5} ($d{=}1024$) \citep{xiaoCPackPackedResources2023}. The resulting embeddings are stored in a FAISS \texttt{IndexFlatIP} index to enable exact maximum inner product search. This setup ensures that any observed gains are attributable to question decomposition and reranking, rather than approximations introduced by approximate nearest neighbor search \citep{douzeFaissLibrary2024,facebookresearchFaissIndexes2024}.

\paragraph{Reranker} We rescore the retrieved passages using the \texttt{bge-reranker-large} cross-encoder \citep{xiaoCPackPackedResources2023}. The model outputs a relevance logit for each query–passage pair. We then sort the passages by their scores and retain the top-$k$ passages, which are appended to the prompt for answer generation.

\paragraph{Generation Model} We generate answers using \texttt{Qwen2.5-32B-Instruct} \citep{qwenteamQwen25PartyFoundation2024,yangQwen2TechnicalReport2024}, operating in bfloat16 precision. We use maximum sequence length of 512 tokens.


\paragraph{Software} In our implementations, we use \texttt{LangChain} \citep{langchainLangChain2025}, \texttt{Huggingface Transformers} \citep{wolfHuggingFacesTransformersStateoftheart2020}, and \texttt{faiss-cpu} \citep{yamaguchiFaisscpuLibraryEfficient2025}. All our experiments are executed on NVIDIA A100 GPUs with 80GB of memory.

\subsection{Hyperparameters}

We use the following hyperparameters across all experiments: the number of retrieved passages is fixed at \( k = 10 \) for all datasets, consistent with the official evaluation settings of both HotpotQA and MultiHop-RAG. Both sub-query generation and answer synthesis are performed with a sampling temperature of 0.8; and we apply nucleus sampling with \( \text{Top-}p = 0.8 \).

\section{Results} \label{sec:results}
\subsection{MultiHop-RAG}
\label{sec:multihop-results}

\begin{table*}[ht]
\centering
\begin{tabular}{lcccc}
\toprule
\textbf{System} & \textbf{Hits@4} & \textbf{Hits@10} & \textbf{MAP@10} & \textbf{MRR@10} \\
\midrule
\texttt{text-ada-002} (+ \textsc{rr})$^{\dagger}$ & 0.616 & 0.706 & 0.463 & 0.548 \\
\texttt{voyage-02} (+ \textsc{rr})$^{\dagger}$    & 0.663 & 0.747 & \textbf{0.480} & 0.586 \\
\midrule
Naive RAG & 0.611 & 0.781 & 0.217 & 0.464 \\
+ \textsc{qd} & 0.655 & 0.810 & 0.238 & 0.498 \\
+ \textsc{rr} & 0.687 & 0.781 & 0.274 & 0.574 \\
+ \textsc{qd+rr} \textit{(ours)} & \textbf{0.763} & \textbf{0.872} & 0.322 & \textbf{0.635} \\
\bottomrule
\end{tabular}
\caption{Retrieval performance on the MultiHop-RAG \textit{eval} split. $^{\dagger}$: We report the best baselines from \citet{tangMultiHopRAGBenchmarkingRetrievalAugmented2024}, including \texttt{text-ada-002} and \texttt{voyage-002} models with reranking.}
\label{tab:multihoprag-main}
\end{table*}
\footnotetext[1]{Results taken from \citet{tangMultiHopRAGBenchmarkingRetrievalAugmented2024}.}

We present retrieval results on the MultiHop-RAG dataset in Table~\ref{tab:multihoprag-main}. Question decomposition (\textsc{qd}) and reranking (\textsc{rr}) individually improve recall-oriented metrics: \textsc{qd} yields +4.4 percentage points on Hits@4 and +2.9 on Hits@10, while \textsc{rr} achieves a +7.6 point gain on Hits@4. Reranking also substantially improves MAP@10 and MRR@10. Our proposed pipeline, which combines both modules (\textsc{qd+rr}), achieves the strongest results overall, reaching 87.2\% Hits@10 and 0.635 MRR@10.

For comparison, the strongest configurations in the original MultiHop-RAG paper \citep{tangMultiHopRAGBenchmarkingRetrievalAugmented2024}, which use \texttt{text-ada-002} \citep{openaiNewImprovedEmbedding2022} and \texttt{voyage-02} \citep{voyageaiinnovationsinc.VoyageAIHome2024} embeddings with \texttt{bge-reranker-large} reranker. Despite using a smaller embedding model, we demonstrate strong improvements over the reported 74.7\% Hits@10 and 0.586 MRR@10. Our \textsc{qd+rr} thus improves Hits@10 by 16.5\% and MRR@10 by 8.4\%. However, we also notice that our approach falls short on MAP@10.

Interestingly, despite the larger retrieval pool from decomposition, MAP@10 also increases (0.322 vs. 0.274 in \textsc{rr}), suggesting that reranking not only filters noise but leverages the broader context to prioritize relevant passages. These findings reinforce the complementary strengths of QD and reranking: decomposition expands coverage, and reranking restores precision.

\begin{table}[t]
\centering
\setlength{\tabcolsep}{4pt}
\begin{tabular}{lcccc}
\toprule
\textbf{System} & \textbf{EM} & \textbf{F$_1$} & \textbf{P} & \textbf{R}\\
\midrule
Naive RAG           & 25.4 & 31.3 & 33.1 & 31.2 \\
\textsc{qd}         & 26.1 & 32.3 & 34.3 & 32.0 \\
\textsc{rr}         & 26.4 & 32.9 & 35.0 & 32.7 \\
\textsc{qd+rr}      & \textbf{28.1} & \textbf{35.0} & \textbf{37.1} & \textbf{34.8} \\
\midrule
\multicolumn{5}{c}{\emph{supporting-fact metrics}}\\
\midrule
Naive RAG           & 18.4 & 12.0 & 42.8 \\
\textsc{qd}         & 17.0 & 10.6 & 44.1 \\
\textsc{rr}         & \textbf{19.6} & \textbf{12.9} & 44.9 \\
\textsc{qd+rr}      & 17.9 & 11.2 & \textbf{46.8} \\
\midrule
\multicolumn{5}{c}{\emph{joint metrics}}\\
\midrule
Naive RAG           & 8.7 & 5.9 & 20.2 \\
\textsc{qd}         & 8.0 & 5.2 & 20.7 \\
\textsc{rr}         & \textbf{9.5} & \textbf{6.4} & 21.4 \\
\textsc{qd+rr}      & 8.9 & 5.8 & \textbf{23.1} \\
\bottomrule
\end{tabular}
\caption{HotpotQA \textit{dev} results.  Upper block: answer metrics; middle: supporting-fact metrics; lower: joint metrics.}
\label{tab:hotpotqa-main}
\end{table}

\subsection{HotpotQA} \label{sec:hotpot-results}


Table~\ref{tab:hotpotqa-main} presents answer-level, supporting-fact, and joint metrics on the \textit{dev} split of HotpotQA.\footnote{The official test set is hidden; as we do not train new models, we follow standard practice and evaluate on the \textit{dev} set.} Applying question decomposition (\textsc{qd}) alone yields only marginal improvements over the naive RAG baseline, with answer F$_1$ increasing from 31.3 to 32.3 and EM from 25.4 to 26.1. Reranking (\textsc{rr}) leads to stronger gains (F$_1$: 32.9, EM: 26.4), demonstrating its effectiveness in improving retrieval relevance. The combined system (\textsc{qd+rr}) achieves the best overall results, with the highest answer EM (28.1), F$_1$ (35.0), precision (37.1), and recall (34.8), indicating that improved coverage and ranking together lead to better evidence-grounded answers.

For supporting-fact metrics, \textsc{qd+rr} achieves the highest precision (46.8), despite having lower EM (17.9) and F$_1$ (11.2) compared to \textsc{rr}, which achieves the highest supporting-fact EM (19.6) and F$_1$ (12.9). Interestingly, \textsc{qd+rr} achieves the highest supporting-fact and joint precision (46.8 and 23.1, respectively), even though decomposition typically expands the retrieval pool and might be expected to reduce precision. This suggests that reranking effectively filters out less relevant candidates, even when starting from a broader and potentially noisier set. Moreover, the results indicate that decomposed sub-queries may surface complementary evidence that, after reranking, leads to more complete and better-aligned evidence sets. In some cases, a single document may contain answers to multiple sub-parts of a complex query, allowing the system to retrieve multi-hop evidence more efficiently than anticipated. These findings highlight the strength of combining decomposition with reranking: the former improves coverage, while the latter restores precision.



\subsection{Ablation: subqueries generated vs.\ gold evidences}
\label{sec:ablations}


Table~\ref{tab:subq-dist} compares the number of gold evidence sentences per query with the number of subqueries produced by the question decomposition module. We instruct the LLM to generate at most 5 subqueries per query in order to keep our experiments strictly zero-shot. Most questions require only two or three supporting facts (e.g., 67.4\% of HotpotQA have two), yet the LLM almost always generates exactly five subqueries (93.3\% on MultiHop-RAG, 98.6\% on HotpotQA), matching the prompt limit. However, we note that allowing variable-size decomposition could better align with actual evidence needs.

\begin{table}[t]
\centering
\setlength{\tabcolsep}{3pt}
\begin{tabular}{@{}l ccc ccc@{}}
\toprule
 & \multicolumn{3}{c}{\textbf{Gold evidences}} &
   \multicolumn{3}{c}{\textbf{Subqueries}}\\
\cmidrule(lr){2-4}\cmidrule(lr){5-7}
\textbf{Dataset} & 2 & 3 & $\ge$4 & 3 & 4 & 5\\
\midrule
MultiHop-RAG & 42.2 & 30.4 & 15.6 & 0.2 & 5.4 & 93.3\\
HotpotQA     & 67.4 & 24.0 &  8.6 & 0.0 & 0.5 & 98.6\\
\bottomrule
\end{tabular}
\caption{Distribution of required gold evidences vs.\ sub-queries generated by QD. Rows sum to 100 \%; buckets $<\!1\%$ are omitted.}
\label{tab:subq-dist}
\end{table}


\noindent\textbf{Correlation analysis.} Both Pearson and Spearman coefficients are near zero (Table~\ref{tab:subq-corr}), indicating no correlation relationship between the number of sub-queries. This suggests that the LLM does not aim to predict the number of reasoning steps (or “hops”), but instead produces a diverse set of focused subqueries. Importantly, our goal was not to mirror the gold evidence count, but to ensure broad coverage through over-complete decomposition, increasing the chance of retrieving all relevant evidence. The near-zero correlation scores suggest the model applies a fixed subquery “budget” defined by the prompt, rather than adapting to question complexity.

\begin{figure}[t]
\centering
\includegraphics[width=\linewidth]{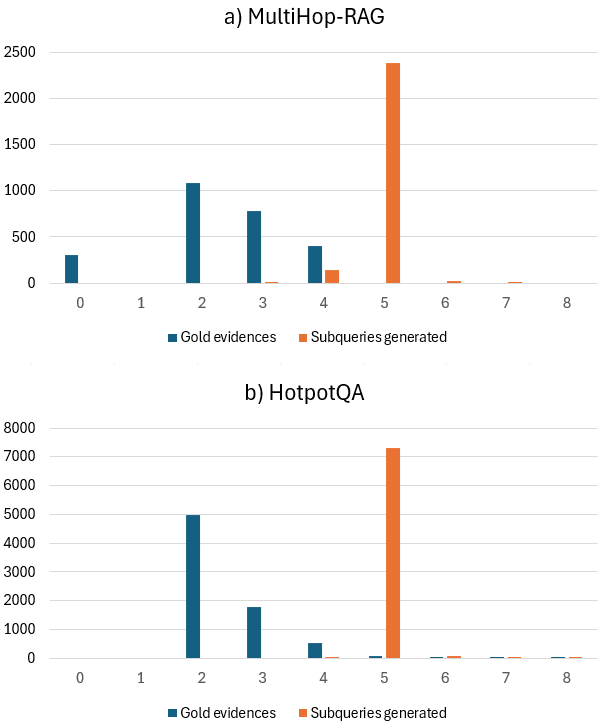}
\caption{Absolute counts of gold evidences (blue) vs.\ subqueries generated (orange).  Left: MultiHop-RAG; right: HotpotQA.}
\label{fig:evidence-vs-subqueries}
\end{figure}

\subsection{Efficiency}
\label{sec:efficiency}

Table~\ref{tab:timings} reports end-to-end retrieval latency (excluding generation) for 250 MultiHop-RAG queries. While Naive RAG is extremely fast (0.03s/query), adding reranking (\textsc{rr}) increases latency substantially to 0.88s/query due to the cost of scoring and sorting candidate passages with a cross-encoder. The overhead of question decomposition (\textsc{qd}) is 16.7s/query. This is primarily due to the additional LLM inference required to generate subqueries. When combined, the full \textsc{qd+rr} system reaches 18.9s/query, thus slower than the simple naive RAG baseline. However, once decomposed, subqueries can be reused (e.g., through caching) so that the latency remains identical to the baseline. A practical implementation is trivial: keep a small key-value store whose key is the raw user query and whose value is the list of generated sub-queries; on a cache hit the expensive QD LLM call is skipped entirely. These results highlight a key tradeoff: while \textsc{qd+rr} achieves the best retrieval quality (Section~\ref{sec:multihop-results}), it does so at the cost of increased latency.

\begin{table}[t]
\centering
\begin{tabular}{lcc}
\toprule
\textbf{System} & \textbf{Total (s)} & \textbf{Per-query (s)}\\
\midrule
Naive RAG      & \phantom{0}7.9 & 0.03 \\
\textsc{rr}    & 219.8 & 0.88 \\
\textsc{qd}    & 4183.9 & 16.7 \\
\textsc{qd+rr} & 4734.9 & 18.9 \\
\bottomrule
\end{tabular}
\caption{Retrieval wall-clock times on 250 MultiHop-RAG queries.}
\label{tab:timings}
\end{table}



\section{Related Work}
\paragraph{Retrieval-Augmented Generation and Multi-Hop QA.} 
RAG augments LLMs with access to external information at inference time, addressing their inherent limitations in handling up-to-date or specialized knowledge \citep{lewisRetrievalAugmentedGenerationKnowledgeIntensive2020}. RAG has shown promise in knowledge-intensive tasks such as open-domain and multi-hop question answering (QA), where single-document retrieval is often insufficient \citep{yangHotpotQADatasetDiverse2018, joshiTriviaQALargeScale2017}. However, RAG performance heavily depends on the quality of retrieved content---irrelevant or misleading passages can significantly impair answer quality \citep{choImprovingZeroshotReader2023, shiLargeLanguageModels2023, yanCorrectiveRetrievalAugmented2024}.

\paragraph{Question Decomposition for Multi-Hop Retrieval.} 
To better address multi-hop queries that span multiple evidence sources, recent work has explored decomposing complex questions into simpler subqueries \citep{feldmanMultiHopParagraphRetrieval2019, yaoReActSynergizingReasoning2023, faziliGenScoCanQuestion2024, xuSearchintheChainInteractivelyEnhancing2024, shaoEnhancingRetrievalAugmentedLarge2023} using large language models as synthetic data generator \citep{golde-etal-2023-fabricator,li-zhang-2024-planning}. This decomposition strategy allows models to target different aspects of a query independently, thereby facilitating more complete evidence aggregation \citep{pressMeasuringNarrowingCompositionality2023}. However, this approach is not without limitations. One notable issue is the "lost-in-retrieval" problem \citep{zhu2025mitigatinglostinretrievalproblemsretrieval}, where LLMs fail to match the recall performance of specialized models such as those trained for named entity recognition \citep{golde-etal-2024-large}. Further, many of these approaches rely on sequential subquestion resolution, which introduces latency and increases the risk of cascading errors \citep{maviMultihopQuestionAnswering2024}. Alternative techniques involve decomposing queries using specialized models or fine-tuning decomposition modules \citep{minMultihopReadingComprehension2019, srinivasanQUILLQueryIntent2022, zhouLearningDecomposeHypothetical2022, wangRichRAGCraftingRich2024, wuGenDecRobustGenerative2024}, limiting their generality. Our work instead adopts a single-step decomposition approach using general-purpose LLMs without task-specific training, ensuring modularity and ease of integration.

\paragraph{Reranking for Precision Retrieval.} 
Reranking methods further refine the retrieval stage by scoring initially retrieved candidates using more expressive models, typically cross-encoders \citep{nogueiraPassageRerankingBERT2020}. These models evaluate query-document pairs jointly, capturing fine-grained interactions and significantly improving relevance over dual-encoder architectures \citep{reimersSentenceBERTSentenceEmbeddings2019}. Reranking has proven effective in boosting precision for multi-hop and complex QA pipelines \citep{tangMultiHopRAGBenchmarkingRetrievalAugmented2024}. Our approach leverages cross-encoder reranking in conjunction with question decomposition, which together enhance both document coverage and ranking quality.

\paragraph{Complementary Approaches.}
A range of complementary strategies has been proposed to optimize retrieval for complex queries, including adaptive retrieval \citep{jeongAdaptiveRAGLearningAdapt2024}, corrective reranking \citep{yanCorrectiveRetrievalAugmented2024}, and self-reflective generation \citep{asaiSelfRAGLearningRetrieve2023}. Techniques such as hypothetical document embeddings (HyDE) \citep{gaoPreciseZeroShotDense2022} and query rewriting \citep{chanRQRAGLearningRefine2024, maQueryRewritingRetrievalAugmented2023} focus on improving the retrieval query itself. While promising, many of these methods involve non-trivial training or model customization. In contrast, our method is lightweight, model-agnostic, and easily deployable within existing RAG architectures.

\section{Conclusion}
This study examined how LLM-based question decomposition (QD) and cross-encoder reranking influence retrieval-augmented generation for complex and multi-hop question answering. Across four system variants and two datasets, the combination of QD and reranking provided the largest gains, increasing retrieval and answer correctness, without requiring extra training or domain-specific tuning. Splitting a query into focused sub-queries broadened evidence coverage, while the reranker promoted the most relevant passages, yielding improvements on benchmark datasets.\\
But the approach is not without downsides. If a query is already precise, decomposition can introduce noise, and reranking cannot remove every irrelevant passage. Both modules also add computation, which may be prohibitive in low-latency scenarios. Performance further depends on the quality of the LLM used for sub-query generation and on an appropriate choice of reranker.

\noindent\textbf{Future work.}  
Employing QD only when a query is predicted to need multi-hop reasoning could preserve most benefits while cutting overhead. The incorporation of both QD and reranking inevitably increases computational overhead, which can be a limitation in low-latency, real-time deployments. Future work could therefore focus on efficiency-oriented variants, e.g.\ swapping in smaller instruction models for QD or using lightweight rerankers, to keep response times low without sacrificing accuracy. Additional gains may come from testing alternative LLMs, rerankers and prompts, and from tuning the number of sub-queries and retrieved passages. Additionally, human studies and domain-specific evaluations can deepen our understanding of real-world impact and clarify how generated sub-queries relate to required evidence.

\section*{Limitations}

While our approach improves multi-hop retrieval quality, it has several limitations that warrant further attention.

\noindent\textbf{Single-hop and adverse cases.} Question decomposition can be counterproductive when the original query is already specific. In such cases, subqueries may introduce noise or distract from the original intent. In rare instances, none of the generated subqueries retrieve stronger evidence than the original query alone.

\noindent\textbf{Prompt and model sensitivity.} The quality of subqueries is sensitive to both the prompt design and the underlying LLM. This dependence may require prompt tuning or model selection when adapting the method to new domains or languages, potentially limiting generalization.

\noindent\textbf{Computational overhead.} As discussed in §\ref{sec:efficiency}, generating \( M \) subqueries and reranking \( M \times k \) candidate passages substantially increases latency and GPU requirements. This motivates future work on more efficient decomposition strategies, such as lightweight LLMs, retrieval-aware early stopping, or subquery caching.

\noindent\textbf{Pipeline complexity.}
Our design adds two separate modules to the standard RAG stack. Although both are plug-and-play, and rerankers are already commonly used in RAG pipelines \citep{saxenaRankingFreeRAG2025}, every extra component increases engineering overhead, latency, and potential points of failure.

\noindent\textbf{Reranker and domain dependence.} The observed gains rely on a strong, domain-aligned cross-encoder reranker. When the reranker is mismatched with the retrieval or task domain, the benefits of decomposition may diminish or vanish entirely.

\noindent\textbf{Lack of iterative retrieval.} Our pipeline operates in a single-shot manner: subqueries are generated once and not updated based on retrieved evidence. This limits its ability to support adaptive multi-step reasoning, which might be necessary for more complex tasks.

\section*{Acknowledgments}
We thank all reviewers for their valuable comments. Jonas Golde is supported by the Bundesministerium für Bildung und Forschung (BMBF) as part of the project ``FewTuRe'' (project number 01IS24020). Alan Akbik is supported by the Deutsche Forschungsgemeinschaft (DFG, German Research Foundation) under Emmy Noether grant ``Eidetic Representations of Natural Language'' (project number 448414230) and under Germany’s Excellence Strategy ``Science of Intelligence'' (EXC 2002/1, project number 390523135).


\bibliography{citations}

\appendix
\section{Additional Ablation Results}
\label{app:corr}
\begin{table}[htbp]
\centering
\small
\begin{tabular}{lcc}
\toprule
\textbf{Dataset} & \textbf{Pearson (p)} & \textbf{Spearman (p)}\\
\midrule
MultiHop-RAG & $-0.022\;(0.27)$ & $-0.007\;(0.71)$\\
HotpotQA     &  $0.017\;(0.15)$ &  $0.012\;(0.32)$\\
\bottomrule
\end{tabular}
\caption{Correlation between the number of sub-queries and the number of gold evidences per query.}
\label{tab:subq-corr}
\end{table}

\end{document}